\documentclass[11pt]{article}

\usepackage[final]{acl}
\usepackage{times}
\usepackage{latexsym}

\usepackage[LAE,LFE]{fontenc}
\usepackage[T1]{fontenc}
\usepackage{microtype}
\usepackage{inconsolata}
\usepackage{graphicx}
\usepackage{subfig}
\usepackage{adjustbox}
\usepackage{booktabs}
\usepackage{multirow}
\usepackage[table,xcdraw]{xcolor}
\usepackage{listings}
\usepackage{comment}
\usepackage{ifthen}
\usepackage{threeparttable}
\usepackage{siunitx}
\usepackage{CJKutf8}

\usepackage{tabularx}
\usepackage{makecell}

\usepackage{polyglossia}
\setmainlanguage{english}
\setotherlanguage{arabic}
\setotherlanguage{chinese}

\newfontfamily\chinesefont{gbsn00lp.ttf}[Script=CJK]

\newfontfamily\arabicfont[Script=Arabic]{Amiri-Regular.ttf}
\newfontfamily\arabicfontsf[Script=Arabic]{Amiri-Regular.ttf}
\newfontfamily\arabicfonttt[Script=Arabic]{Amiri-Regular.ttf}

\usepackage{pifont}

\newcommand{\ar}[1]{\textarabic{#1}}

\definecolor{codegreen}{rgb}{0,0.6,0}
\definecolor{codegray}{rgb}{0.5,0.5,0.5}
\definecolor{codepurple}{rgb}{0.58,0,0.82}
\definecolor{backcolour}{rgb}{0.95,0.95,0.92}

\lstdefinestyle{mystyle}{
    backgroundcolor=\color{backcolour},
    commentstyle=\color{codegreen},
    keywordstyle=\color{blue},
    numberstyle=\tiny\color{codegray},
    stringstyle=\color{codepurple},
    basicstyle=\footnotesize\ttfamily,
    breakatwhitespace=false,
    breaklines=true,
    captionpos=b,
    keepspaces=true,
    numbers=left,
    numbersep=5pt,
    showspaces=false,
    showstringspaces=false,
    showtabs=false,
    tabsize=2
}

\newboolean{showcomments}
\setboolean{showcomments}{true}
\ifdefined\notodocomments
  \setboolean{showcomments}{false}
\fi
\ifthenelse{\boolean{showcomments}}
 { \newcommand{\mynote}[2]{
      \fbox{\bfseries\sffamily\scriptsize#1}
        {\small$\blacktriangleright$\textsf{\emph{#2}}$\blacktriangleleft$}}}
        { \newcommand{\mynote}[2]{}}

\title{
\raisebox{-0.25\height}{\includegraphics[height=3em]{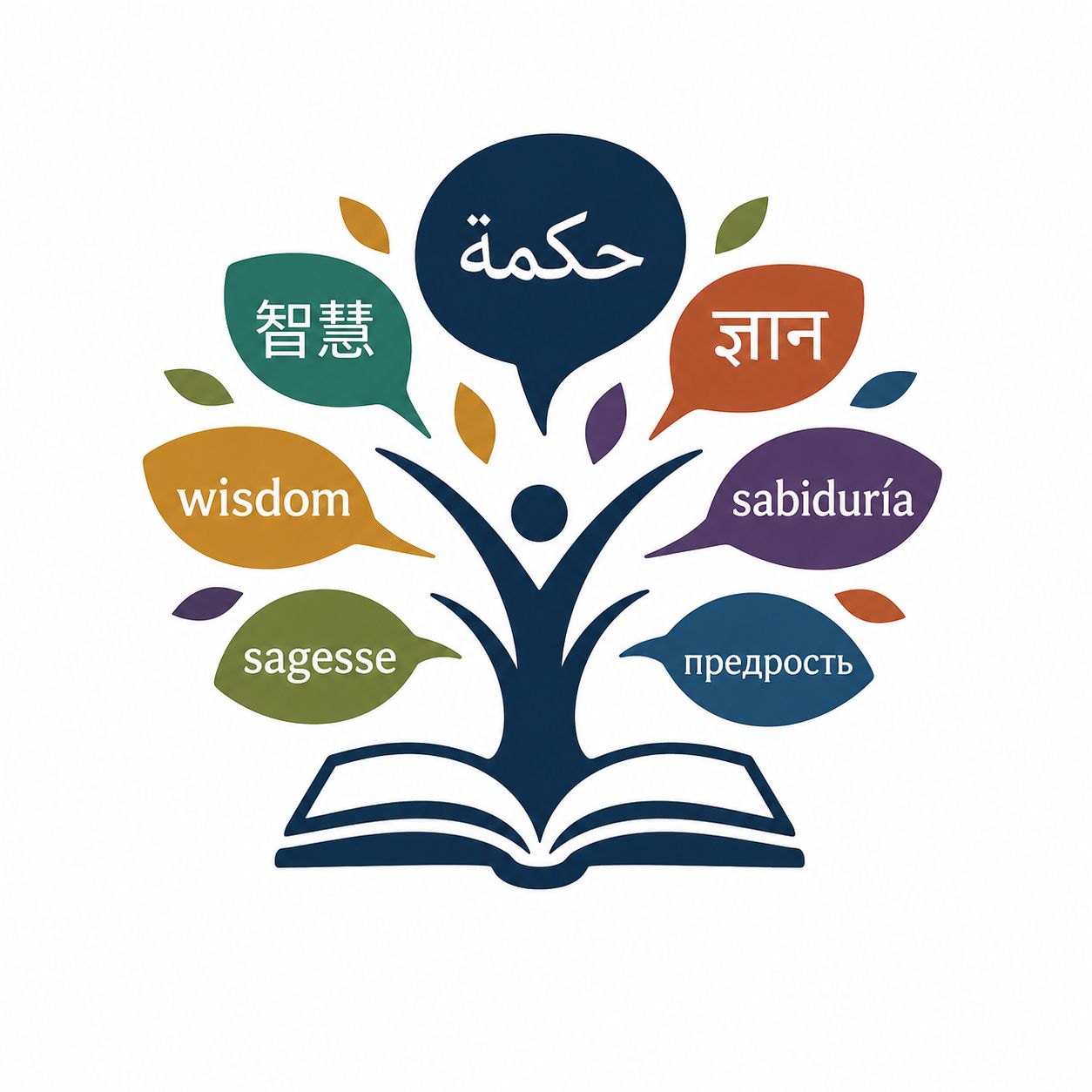}}
\hspace{0.3em}
\textsc{WISDOM}: A Cross-Regional Multilingual Benchmark for Proverbs and Figurative Language
}

\author{
Salmane Chafik\textsuperscript{1},
Salima Lamsiyah\textsuperscript{2},
Saad Ezzini\textsuperscript{3},
Ayoub Nainia\textsuperscript{4},
Christophe Friezas Gonçalves\textsuperscript{2},\\
Abrar Alotaibi\textsuperscript{3},
Maram Alharbi\textsuperscript{5},
Alicia Picazo-Izquierdo\textsuperscript{6},
Nouran Khallaf\textsuperscript{5},
Xiaojing Zhao\textsuperscript{7},\\
Shadi Abudalfa\textsuperscript{3},
Nadia Ghezaiel Hammouda\textsuperscript{8},
Aadil Gani Ganie\textsuperscript{3},
Hamzah Luqman\textsuperscript{3}
\\
\\
\textsuperscript{1} Mohammed VI Polytechnic University, Morocco
\quad
\textsuperscript{2} University of Luxembourg, Luxembourg
\\
\textsuperscript{3} King Fahd University of Petroleum and Minerals, Saudi Arabia
\\
\textsuperscript{4} Sorbonne University, France
\quad
\textsuperscript{5} Lancaster University, UK
\\
\textsuperscript{6} University of Alicante, Spain
\quad
\textsuperscript{7} Hong Kong Polytechnic University, Hong Kong
\\
\textsuperscript{8} University of Hail, Saudi Arabia
\quad
}

\makeatletter
\AtBeginDocument{
  \let\@oldmaketitle\@maketitle
  \renewcommand{\@maketitle}{
    \@oldmaketitle
    \vspace{45pt}
    \begin{center}
      \includegraphics[width=\linewidth]{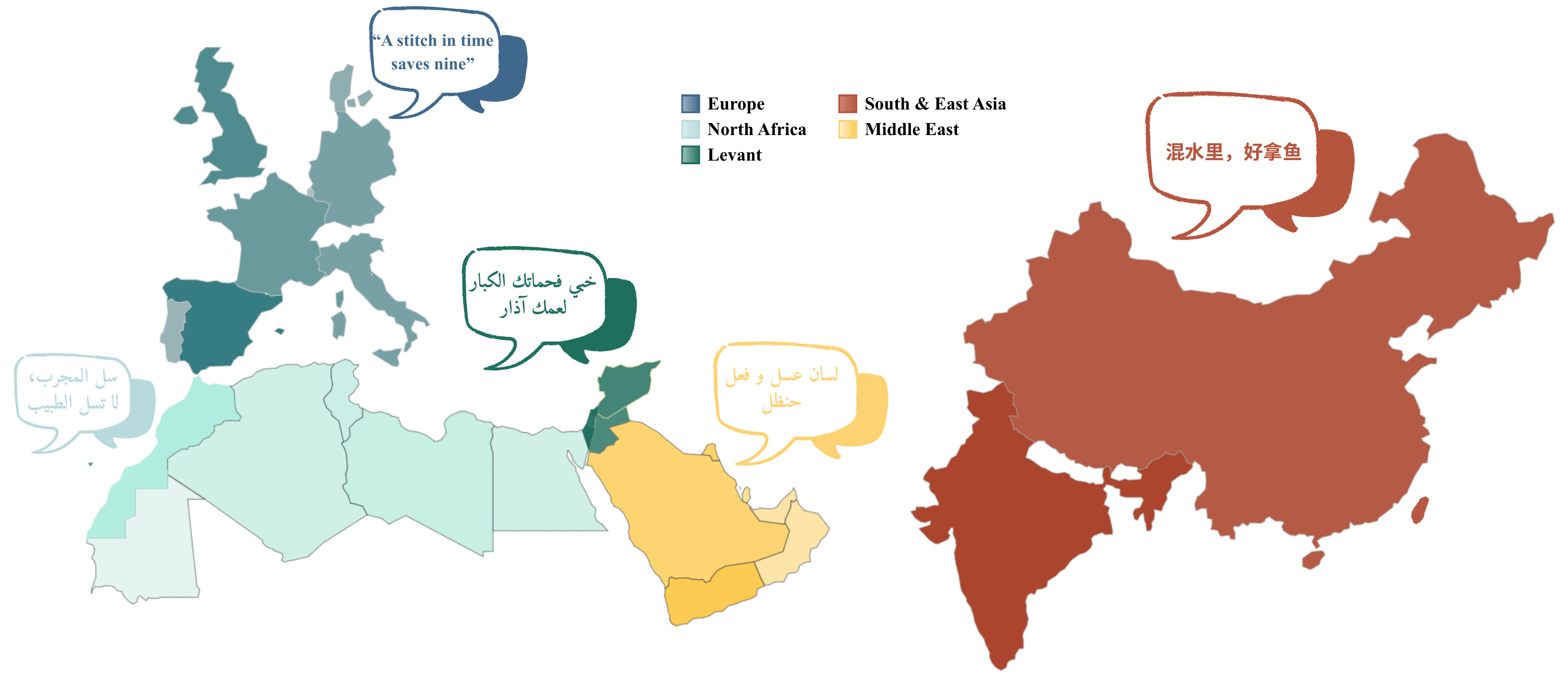}

      \small
      \textbf{Figure 1:} Geographical distribution and linguistic examples
      of regional proverbs across Europe, North Africa, Levant,
      Arabian Peninsula, and South \& East Asia.
    \end{center}
    \vspace{10pt}
  }
}
\makeatother

\begin{document}
\maketitle

\begin{abstract}
\textsc{WISDOM} is a multilingual benchmark for evaluating whether large language models can interpret culturally grounded figurative language rather than merely translate it. The dataset contains 39,838 proverbs across 12 distinct languages and 18 regional dialects/varieties (totaling 30 varieties), with native-speaker annotations that separate original proverb wording from pragmatic meaning, English explanations, semantic categories, and cross-lingual proverb equivalents. \textsc{WISDOM} is organized into two complementary resources: \textsc{WISDOMBase}, which supports proverb explanation and semantic classification, and \textsc{WISDOMBridge}, which links proverbs across cultures through validated English-equivalent proverbs. We evaluate seven proprietary and open-source LLMs on three tasks: proverb explanation, semantic category prediction, and cross-cultural proverb linking. Results show that proprietary systems perform substantially better than open-source baselines, but all models struggle with dialectal, low-resource, and culturally specific proverbs. Classification and proverb linking remain especially challenging, revealing a persistent gap between surface multilingual fluency and robust cultural-pragmatic understanding.
\end{abstract}

\section{Introduction} \label{introduction} 

Large language models (LLMs) have made rapid progress on multilingual language understanding, yet much of their evaluation still emphasizes literal meaning, factual knowledge, or decontextualized reasoning. Cultural knowledge is harder to measure: it is often implicit, locally grounded, and expressed through forms whose surface meanings are deliberately indirect~\citep{kabra-etal-2023-multi,liu-etal-2024-multilingual}. Proverbs are a particularly dense test case: a short expression may encode social norms, ethical judgments, intergenerational experience, humor, fatalism, or advice about everyday life~\citep{liu-etal-2024-multilingual,magdy-etal-2025-jawaher}. Understanding a proverb therefore requires more than translating its words; it requires recovering the pragmatic lesson behind culturally specific imagery~\citep{magdy-etal-2025-jawaher,wang2025proverbs}.

This challenge is central for multilingual NLP: culturally specific proverb imagery may be transparent within one community but opaque elsewhere, and similar lessons can be expressed through very different forms across languages. While commonsense and cultural benchmarks evaluate social reasoning and everyday knowledge \citep{sap-etal-2019-social,huang2019cosmos,myung2024blend,ayash2025saudiculture,sadallah-etal-2025-commonsense,almheiri2025cross}, culturally grounded figurative language remains underrepresented in large-scale multilingual evaluation.


Prior work has introduced resources for metaphor, idioms, figurative explanation, proverb-in-context interpretation, and multilingual figurative language evaluation \citep{ozbal-etal-2016-prometheus,ghosh-srivastava-2022-epic,chakrabarty-etal-2022-flute,liu-etal-2024-multilingual,park-etal-2025-fluid}, with recent studies focusing on language-specific, low-resource, and translation-oriented proverb benchmarks \citep{magdy-etal-2025-jawaher,thapa-etal-2025-probing,wang2025proverbs,mensa2025easy,olusanya2025owe}. However, most resources cover limited tasks, regions, or languages, and rarely combine native-speaker explanations, semantic categories, and explicit cross-lingual proverb equivalence at broad multilingual scale.

We introduce \textsc{WISDOM} (Cross-Regional Multilingual Benchmark for Proverbs and Figurative Language), a human-annotated benchmark of 39,838 proverbs across 12 distinct languages and 18 regional dialects/varieties (totaling 30 varieties). Each proverb preserves its original form and includes native-speaker explanations in the original language and English. \textsc{WISDOMBase} supports proverb understanding and semantic classification, while \textsc{WISDOMBridge} supports cross-cultural retrieval through validated English-equivalent proverb links.

\textsc{WISDOM} enables three evaluations of multilingual figurative competence: proverb interpretation, semantic abstraction, and cross-cultural alignment. Experiments with current LLMs show that proprietary models outperform open-source baselines, but all models struggle with dialectal Arabic, South and East Asian languages, culturally specific proverbs, semantic classification, and exact cross-cultural proverb linking.

The main contributions of this work are as follows:
\begin{itemize}
    \item We introduce \textsc{WISDOM}\footnote{The dataset is available at \url{https://huggingface.co/datasets/salmane11/WISDOM}.}, a large-scale multilingual benchmark for culturally grounded figurative language, covering 39,838 proverbs across 12 distinct languages and 18 regional dialects/varieties (totaling 30 varieties), with annotations produced and validated by native speakers.
    \item We release \textsc{WISDOMBase}, which pairs each proverb with original-language and English pragmatic explanations, semantic macro-categories, and optional fine-grained subcategories, enabling evaluation of proverb interpretation and semantic abstraction.
    \item We release \textsc{WISDOMBridge}, a cross-lingual equivalence resource that aligns proverbs to validated English-equivalent proverbs and supports source--target proverb retrieval by pragmatic meaning rather than surface translation.
    \item We propose an evaluation framework for LLM proverb understanding, semantic classification, and cross-cultural proverb retrieval, enabling analysis across languages, regions, and cultural communities.
\end{itemize}

\section{Related Work} \label{relatedwork}

\paragraph{Commonsense and Cultural Evaluation.} Benchmarks for commonsense reasoning have long been used to test whether models can infer unstated social, causal, and everyday knowledge. SocialIQA evaluates reasoning about social interactions \citep{sap-etal-2019-social}, while CosmosQA targets contextual commonsense in reading comprehension \citep{huang2019cosmos}. These benchmarks are valuable, but they are not designed to capture culturally specific figurative meaning. More recent work has argued that LLMs should be evaluated on culturally diverse knowledge rather than only on universal or English-centric assumptions. BLEND evaluates everyday knowledge across diverse cultures and languages \citep{myung2024blend}; SaudiCulture targets cultural competence within Saudi Arabia \citep{ayash2025saudiculture}; and Arabic-focused work has examined commonsense, dialectal variation, and cross-cultural transfer in the Arab world \citep{sadallah-etal-2025-commonsense,alwajih2025palm,keleg2025revisiting,almheiri2025cross}. \textsc{WISDOM} follows this line of work but focuses on proverbs as cultural artifacts. Unlike factual cultural questions, proverbs often express values indirectly. A model must infer a general lesson from local imagery, then compare that lesson to meanings expressed in other languages. This makes proverbs useful for testing not only cultural knowledge, but also abstraction, pragmatic interpretation, and cross-lingual semantic alignment.

\paragraph{Figurative Language, Idioms, and Proverbs.} Figurative language has been studied through metaphor detection, idiom identification, textual explanation, and contextual interpretation. Early cross-lingual metaphor work showed that figurative meaning can share semantic features across languages while remaining difficult to transfer directly \citep{tsvetkov-etal-2013-cross}. MAGPIE provides a large corpus of potentially idiomatic expressions \citep{haagsma-etal-2020-magpie}; FLUTE evaluates figurative language understanding through textual explanations \citep{chakrabarty-etal-2022-flute}; and recent surveys document the diversity and limitations of idiom datasets for computational and psycholinguistic research \citep{flor-etal-2025-survey,matheny2025nlp}. Because many figurative expressions are conventionalized and widely repeated, recent work also asks whether LLM success reflects memorization rather than genuine reasoning \citep{kim-etal-2025-memorization}. Work on multilingual figurative language has further shown that model performance varies substantially across languages and expression types \citep{liu-etal-2024-multilingual,park-etal-2025-fluid,biyik-etal-2026-semantic}. Proverbs have received increasing attention as a testbed for abstract language understanding. PROMETHEUS annotates proverbs with metaphors \citep{ozbal-etal-2016-prometheus}, and ePiC uses proverbs in context to evaluate abstract understanding \citep{ghosh-srivastava-2022-epic}. Other recent resources focus on proverb translation, culturally diverse sayings, low-resource language proverbs, and language-specific collections such as Arabic, Greek, Italian, and Yoruba proverb datasets \citep{wang2025proverbs,liu-etal-2024-multilingual,thapa-etal-2025-probing,magdy-etal-2025-jawaher,pavlopoulos-etal-2024-towards,mensa2025easy,olusanya2025owe}. These works establish proverbs as an important benchmark genre. \textsc{WISDOM} extends this direction by combining large-scale multilingual coverage, native-speaker pragmatic explanations, a shared semantic label inventory, and explicit cross-lingual equivalence links.

\paragraph{Cross-Lingual Alignment and Benchmark Construction.} Multilingual representation learning has enabled strong transfer across languages \citep{conneau-etal-2020-unsupervised}, but cross-lingual transfer is not equivalent to cross-cultural understanding. Figurative expressions can be semantically close while differing in register, connotation, imagery, or use conditions. In some cases, semantic overlap is insufficient for reliable transfer, as shown in recent work on cross-lingual euphemism transfer \citep{biyik-etal-2026-semantic}. Proverb alignment intensifies this problem: the best equivalent is often not a literal translation, but a culturally conventional expression that performs the same pragmatic function. Benchmark construction also requires careful documentation and annotation design. Datasheets for Datasets emphasize transparency about data sources, collection procedures, annotation, and intended use \citep{gebru2021datasheets}, while recent work questions when and how benchmark scores can be trusted \citep{gevers-etal-2025-benchmarks}. \textsc{WISDOM} addresses these concerns by recording source and validation metadata where available, preserving original language forms, separating literal wording from pragmatic explanations, and making cross-lingual alignment decisions explicit through validated English-equivalent proverbs.

A detailed comparison of \textsc{WISDOM} with existing proverb datasets, including their language coverage, size, semantic categories, equivalence annotations, and evaluation tasks, is provided in Appendix \ref{app:related_work}, Table \ref{tab:related-work}.

\section{\textsc{WISDOM} Dataset Construction}

\begin{figure}
    \centering
    \includegraphics[width=1\linewidth]{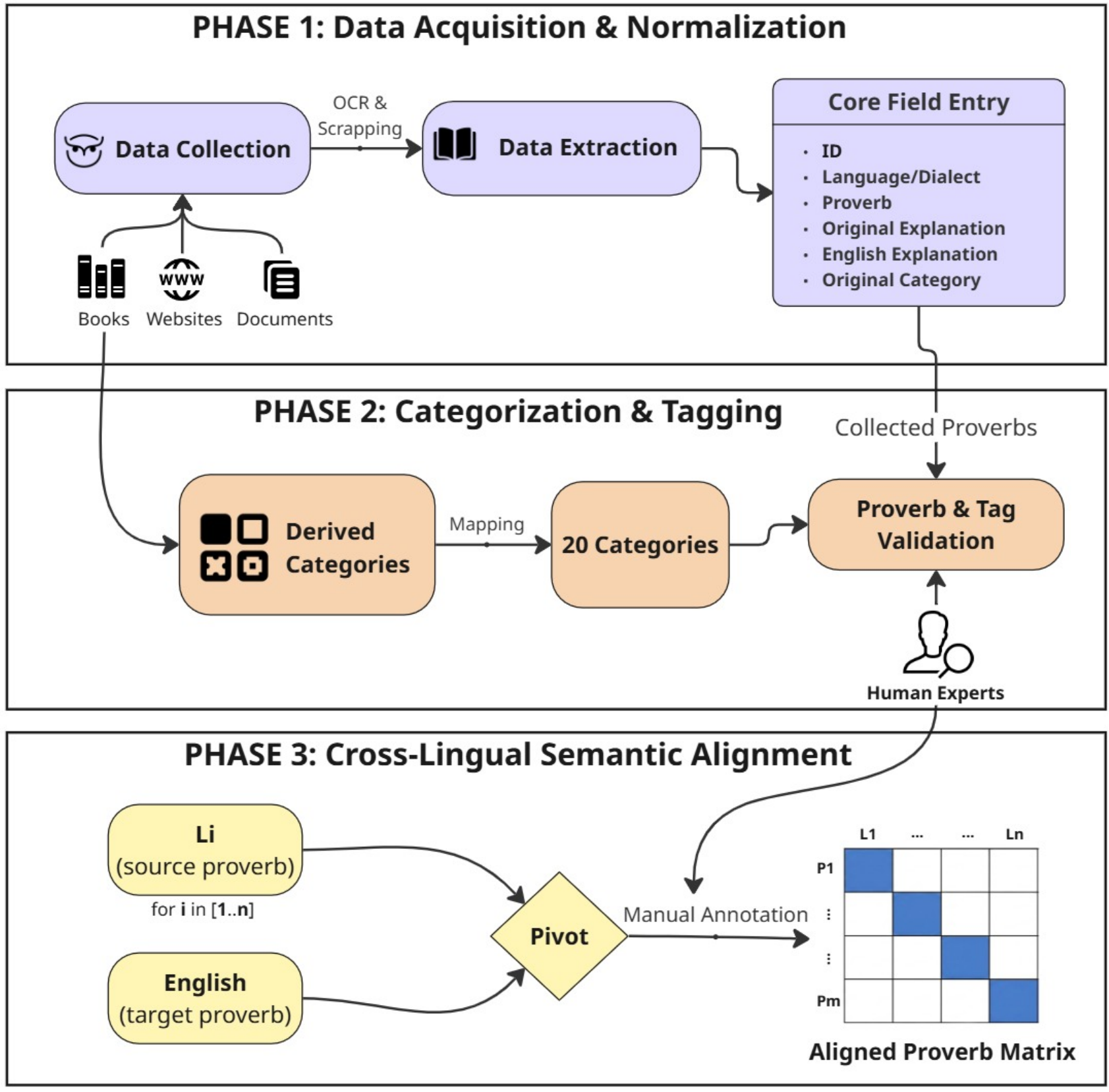}
\caption{Overview of the three-phase \textsc{WISDOM} construction pipeline.}
\label{fig:wisdom-pipeline}
\end{figure}

The process of building \textsc{WISDOM} is illustrated in Figure~\ref{fig:wisdom-pipeline} and described in detail in the following subsections.

\subsection{Design Principles}

\textsc{WISDOM} is built to evaluate whether LLMs can interpret culturally grounded figurative language rather than merely translate it. The annotation protocol is based on one central principle: separate literal wording from pragmatic meaning. Annotators were instructed to preserve the proverb as it appears in the source material or oral tradition, while explaining the lesson, human behavior, or social norm conveyed by the expression. This distinction is especially important for dialectal and regional varieties, where normalizing spelling can erase local form and where literal translation may obscure the intended meaning.

The dataset contains 39,838 proverbs across 30 language and regional varieties. Proverbs were collected from written collections, online resources, and oral tradition where available. Each entry records the macro-language, regional variety when applicable, the original proverb, explanations in the original language and in English, semantic labels, and metadata such as source and annotator validation fields. All annotations were produced or validated by native speakers of the corresponding language or dialect. Details on the language-specific collection process, annotator organization, and source types are provided in Appendix~\ref{app:collection-protocol}.

The project was carried out by a distributed team of native-speaking contributors using a structured coordination process to ensure consistency and steady progress. We held weekly meetings to review annotation progress, discuss recurring issues, and refine the guidelines when needed. Day-to-day communication was managed through a dedicated Slack workspace, while an organized shared drive was used to maintain updated guidelines, data files, progress reports, and versioned resources.

\subsection{\textsc{WISDOMBase}: Proverb Meaning and Semantic Labels}

\textsc{WISDOMBase} is the proverb-level subset. It supports tasks that require understanding a proverb in isolation: generating an explanation, selecting the correct interpretation, or assigning a semantic category. Table~\ref{tab:wisdom-schema} in Appendix~\ref{app:wisdom-schema} summarizes the core fields. For the English explanation, annotators were instructed to remove the figurative scene and state the underlying lesson. For example, instead of explaining a proverb by repeating its animal, food, or kinship imagery, the annotation should describe the general idea: inherited traits, misplaced trust, false pride, the danger of haste, or the value of experience. This makes the explanation useful for evaluating semantic understanding independent of surface translation.

\subsection{Semantic and Pragmatic Categories}

Each proverb receives one primary macro-category. The category inventory was developed through an iterative process that combined bottom-up evidence from the collected sources with cross-lingual harmonization. When source collections already provided thematic labels, we used them as initial guidance; annotators then compared these source-specific categories across languages and finalized a shared inventory of 20 macro-categories. This inventory was designed to cover broad pragmatic functions common in proverb traditions while remaining stable enough for multilingual classification. Table~\ref{tab:wisdom-categories} in Appendix~\ref{app:wisdom-schema} lists the categories used in the annotation. Annotators could also provide an optional subcategory to capture finer distinctions within a macro-category. The category label supports the second evaluation task: testing whether LLMs can abstract away from the literal proverb and identify its pragmatic domain. This is different from topical keyword matching. A proverb mentioning animals, for instance, may belong to Character, Ethics \& Morality or Wisdom, Knowledge \& Experience rather than Nature, Animals \& Environment if the animal image is only a metaphor for human behavior.

\subsection{\textsc{WISDOMBridge}: Cross-Lingual Proverb Equivalence}
\textsc{WISDOMBridge} is the cross-lingual alignment subset. Its goal is to link proverbs that express the same pragmatic meaning across languages, even when their literal imagery is unrelated. The alignment procedure uses English as a practical reference language, not as a claim that English is culturally neutral. Each non-English proverb is directly linked to one validated English proverb when a clear meaning-based equivalent exists. Annotators searched the English proverb inventory and accepted a link only when the target English proverb conveyed the same pragmatic lesson rather than merely sharing words, imagery, or topic.

This procedure produces a pairwise, meaning-centered resource. Cross-language source--target pairs can be derived when two proverbs are linked to the same validated English equivalent, but the annotation itself remains a direct proverb-to-English-proverb decision. The resulting structure supports source-to-target proverb retrieval: given a proverb in one language, a model must infer the underlying lesson and retrieve an equivalent proverb in another language. This task is substantially harder than translation because the correct answer may use different imagery, lexical content, and syntactic form.

\subsection{Dataset Statistics and Coverage}
Table~\ref{tab:wisdom-statistics} summarizes the current composition of \textsc{WISDOM}. The benchmark is organized by language and regional variety, covering Classical Arabic, Modern Standard Arabic, Arabic varieties from the Arabian Peninsula, the Levant, and North Africa, as well as European/USA and South/East Asian subsets. The dataset is intentionally not fully balanced: some proverb traditions are more extensively documented than others, and downsampling would remove cultural and lexical diversity. For classification, where category distributions are imbalanced, we therefore report weighted-average metrics in addition to accuracy.

\begin{figure}
    \centering
    \includegraphics[width=1\linewidth]{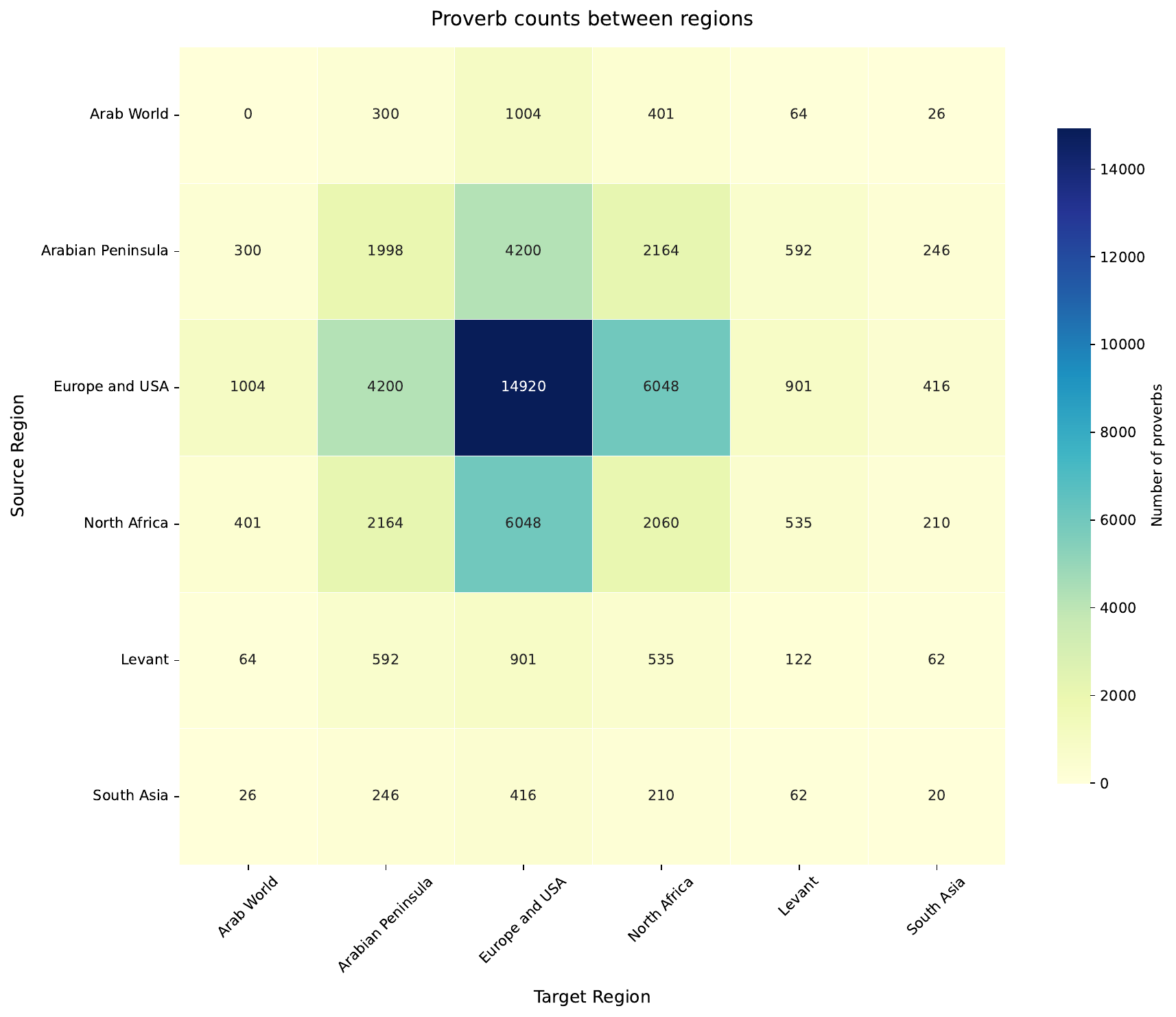}
    \caption{Heatmap showing the number of equivalent proverbs between source and target regions across \textsc{WISDOMBridge}.}
    \label{fig:proverbs-heatmap}
\end{figure}

Table~\ref{tab:wisdom-stats} summarizes the regional composition of \textsc{WISDOMBase}. \textsc{WISDOMBridge} extends this resource by linking proverbs across regions through shared pragmatic meanings rather than literal translations; Figure~\ref{fig:proverbs-heatmap} shows the resulting cross-regional alignment patterns, with dense links within Europe/USA and between Europe/USA and North Africa or the Levant, and sparser links involving South/East Asia and the Arab World.

\begin{table}[h!]
\centering
\scriptsize
\setlength{\tabcolsep}{2pt}
\begin{tabular}{@{}llrr@{}}
\toprule
\textbf{Region} & \textbf{Country/Language} & \textbf{Size} & \textbf{Total} \\
\midrule
\multirow{2}{*}{\textbf{Arab World}}
& Classical Arabic & 132 & \multirow{2}{*}{752} \\
& Modern Standard Arabic & 620 & \\
\midrule
\multirow{8}{*}{\textbf{Arabian Peninsula}}
& Arabian Peninsula (mixed) & 1665 & \multirow{8}{*}{9964} \\
& Bahrain & 530 & \\
& Kuwait & 1513 & \\
& Oman & 103 & \\
& Qatar & 1181 & \\
& Saudi Arabia & 1161 & \\
& United Arab Emirates & 305 & \\
& Yemen & 3506 & \\
\midrule
\multirow{3}{*}{\textbf{Levant}}
& Jordan & 1141 & \multirow{3}{*}{7939} \\
& Palestine & 5650 & \\
& Syria & 1148 & \\
\midrule
\multirow{6}{*}{\textbf{North Africa}}
& Algeria & 1003 & \multirow{6}{*}{7385} \\
& Egypt & 1116 & \\
& Libya & 1284 & \\
& Mauritania & 532 & \\
& Morocco & 2290 & \\
& Tunisia & 1160 & \\
\midrule
\multirow{9}{*}{\textbf{Europe/USA}}
& Catalan & 377 & \multirow{9}{*}{11152} \\
& Danish & 1057 & \\
& English & 1988 & \\
& French & 1577 & \\
& German & 1365 & \\
& Italian & 1360 & \\
& Luxembourgish & 303 & \\
& Portuguese & 624 & \\
& Spanish & 2501 & \\
\midrule
\multirow{3}{*}{\textbf{South/East Asia}}
& Hindi & 747 & \multirow{3}{*}{2646} \\
& Kashmiri & 747 & \\
& Mandarin & 1152 & \\
\midrule
\textbf{Total} & & & \textbf{39838} \\
\bottomrule
\end{tabular}
\caption{Distribution of \textsc{WISDOMBase} entries by region and language or regional variety.}
\label{tab:wisdom-stats}
\end{table}

\subsection{Quality Control and Validation}
Quality control combined native-speaker review, a shared annotation protocol, and automated consistency checks. Contributors verified that each proverb was conventional in the relevant language or dialect, preserved original orthography, and ensured that English explanations captured pragmatic meaning rather than literal translation.

Semantic labels were reviewed against the shared 20-category inventory to ensure that categories reflected the proverb's function rather than surface keywords. Ambiguous cases were discussed during coordination meetings to improve consistency across languages and regional varieties.

For \textsc{WISDOMBridge}, annotators accepted proverb-to-English-equivalent links only when both proverbs expressed the same lesson in normal use. Literal translations, broad topical matches, weak paraphrases, and culturally unnatural equivalents were rejected or flagged for revision. The final pipeline checked for missing fields, duplicate identifiers, empty explanations, invalid categories, and inconsistent English-equivalent links.


\section{Evaluation} 

We evaluate state-of-the-art LLMs on \textsc{WISDOMBase} and \textsc{WISDOMBridge} through three tasks that test complementary aspects of multilingual figurative competence: proverb explanation, semantic classification, and cross-cultural proverb linking.


\subsection{Task Definitions}

\paragraph{\texttt{Proverb Explanation.}}
Given a proverb $P$ written in language or regional variety $L$, a model $M_\theta$ must generate an English explanation $E'$ that captures the proverb's pragmatic meaning:
\begin{center}
$M_\theta(P, L) = E'$
\end{center}
The reference $E$ is the human-written English explanation in \textsc{WISDOMBase}. We compare $E'$ against $E$ using BLEU, semantic similarity, BERTScore, and human evaluation. The prompt is shown in Listing~\ref{lst:task1_prompt}, and the manual evaluation protocol is described in Appendix~\ref{sec:manual-evaluation}.

\paragraph{\texttt{Proverb Classification.}}
Given a proverb $P$ and its language or regional variety $L$, a model must predict the semantic macro-category $C'$ that best represents the proverb's pragmatic function:
\begin{center}
$M_\theta(P, L) = C'$
\end{center}
The gold label $C$ is the human-annotated category from the 20-category inventory in Table~\ref{tab:wisdom-categories}. We evaluate predictions using accuracy and weighted-average precision and F1, which account for imbalanced category distributions. The prompt is shown in Listing~\ref{lst:task2_prompt}.

\paragraph{\texttt{Proverb Linking.}}
Given a source proverb $P_S$ in language $L_S$ and a target language $L_T$, a model must generate a culturally recognized target-language proverb $P_T'$ that conveys the same pragmatic meaning:
\begin{center}
$M_\theta(P_S, L_S, L_T) = P_T'$
\end{center}
The reference $P_T$ is the human-validated equivalent proverb in \textsc{WISDOMBridge}. The output should not be a literal translation of the source proverb or its explanation; it should be a naturally occurring proverb in the target culture when such an equivalent exists. We evaluate $P_T'$ using BLEU, semantic similarity, BERTScore, and Exact Match. The prompt is shown in Listing~\ref{lst:task3_prompt}.

Our three evaluation tasks assess different aspects of cultural and pragmatic understanding. Task 1 (Proverb Explanation) compares the model-generated English explanation with the human-annotated reference explanation in WISDOMBASE, focusing on the proverb's underlying lesson rather than its literal meaning. Task 2 (Semantic Category Classification) evaluates the model's ability to assign each proverb to one of 20 macro-categories based on its pragmatic meaning, using the human-assigned category as the reference. Task 3 (Cross-Cultural Proverb Linking) tests the model's ability to identify a naturally occurring proverb in another language that expresses the same pragmatic meaning, using human-validated equivalents from WISDOMBRIDGE as references. Together, these tasks measure the ability to interpret, categorize, and connect proverbs across languages and cultures.

\subsection{Evaluation Metrics} \label{Subsec:metrics}

We use both automatic and manual metrics. \textbf{BLEU}~\citep{bleu} measures n-gram overlap between model output and reference. \textbf{Semantic similarity} is computed as cosine similarity between sentence embeddings of the prediction and reference using \texttt{intfloat/multilingual-e5-base}\footnote{https://huggingface.co/intfloat/multilingual-e5-base}. \textbf{BERTScore}~\citep{bertscore} aligns contextual token embeddings between prediction and reference. For the explanation task, we use the default model provided by BERTScore, as the outputs are exclusively in English. For the equivalence task, we use the multilingual \texttt{xlm-roberta-base} model to accommodate sentences in multiple languages. \textbf{Exact Match} checks whether the predicted proverb exactly matches the reference string after normalization. For the explanation task, we additionally conduct \textbf{human evaluation}: native speakers rate model explanations on a 1--5 scale according to cultural correctness, specificity, concision, and avoidance of literal translation. These scores are normalized to $[0,1]$ in the reported tables. Additional details about human evaluation are provided in Appendix~\ref{sec:manual-evaluation}.

All reported metrics are normalized to $[0,1]$ except BLEU, which is reported on the conventional 0-100 scale.


\subsection{Baseline Models} \label{subsection:baselines}

We evaluate seven multilingual LLMs covering proprietary and open-source systems. The proprietary models are \textit{GPT-5.4}~\citep{gpt5}, \textit{Gemini 3 Flash}~\citep{gemini3}, \textit{Claude Haiku 4.5}~\citep{haiku45}, and \textit{DeepSeek V4 Flash}~\citep{deepseekv4}. The open-source models are \textit{Qwen3 8B}~\citep{qwen3}, \textit{Llama 3.1 Instruct 8B}~\citep{llama3}, and \textit{Mistral v0.2 Instruct 7B}~\citep{mistral}. These models were selected to compare systems with different parameter scales, deployment settings, and multilingual capabilities.


\subsection{Experimental Setup}

We evaluate the seven baseline models through three experiments.

\paragraph{EX1: Explanation.}
For RQ1, we randomly sample 100 examples from each of the 31 language and regional subsets in \textsc{WISDOMBase}, yielding \textbf{\texttt{Test\_1}} with 3,100 examples. Each example contains a proverb, its language or regional variety, its English explanation, and its semantic category. We prompt each model using Listing~\ref{lst:task1_prompt}. Automatic evaluation uses BLEU, semantic similarity, and BERTScore. We also conduct manual evaluation for all seven models on a stratified subset of 310 examples, denoted as \textbf{\texttt{Test\_1\_sub}}. The manual evaluation protocol is described in Appendix~\ref{sec:manual-evaluation}.


\paragraph{EX2: Classification.}
For RQ2, we use the same \textbf{\texttt{Test\_1}} split and ask each model to assign one of the 20 macro-categories listed in Table~\ref{tab:wisdom-categories}. The prompt is shown in Listing~\ref{lst:task2_prompt}. We report accuracy and weighted-average precision and F1 because category frequencies are imbalanced.


\paragraph{EX3: Linking.}
For RQ3, we construct \textbf{\texttt{Test\_2}} by sampling 20 source-target proverb pairs for each source language from \textsc{WISDOMBridge}. Each example contains the source language, source proverb, target language, and target proverb. Table~\ref{tab:test-2-examples} provides examples. We prompt each model using Listing~\ref{lst:task3_prompt} and evaluate the generated target proverb using BLEU, Exact Match, BERTScore, and semantic similarity.

All experiments were conducted in a zero-shot setting. Further details about the experimental setup and the prompts used are provided in Appendices~\ref{app:implementation-details} and \ref{app:prompts}.

\section{Results and Discussion}
\label{sec:experiments}

We organize the experiments around three research questions. 

\begin{itemize}
    \item  \textbf{RQ1: Can existing LLMs explain the deep meaning of traditional proverbs across languages and cultures?}
This question evaluates whether models can recover the pragmatic message of a proverb rather than restating or translating its surface form.

\item \textbf{RQ2: Can LLMs classify proverbs into culturally meaningful semantic categories?}
This question tests whether models can abstract from the proverb wording to a higher-level conceptual function, such as ethics, social relations, or appearance versus reality.

\item \textbf{RQ3: Can LLMs link cultures through equivalent proverbs?}
This question evaluates whether models can generate a culturally recognized proverb in a target language that conveys the same pragmatic meaning as a source-language proverb.
\end{itemize}

Model-level results are reported in the main text, while language- and region-level results are provided in Appendix~\ref{app:additional-results}.

\subsection{RQ1: Proverb Explanation}

Table~\ref{tab:task1-models} shows a substantial gap between proprietary LLMs and open-source models in explaining proverb meanings. Gemini 3 Flash achieves the best manual evaluation score (0.79), with a semantic similarity score of 0.47 and a BERTScore of 0.87. This suggests that it explains many proverbs effectively, although some outputs still lack cultural depth. In contrast, the strongest open-source model, Qwen3, reaches only 0.31 in manual evaluation, indicating limited ability to explain proverbs accurately.

Human evaluation shows that open-source models often fail to capture intended meanings, generate off-context explanations, or violate prompt constraints by reproducing the proverb in the original language. Proprietary models perform better overall, but still struggle with culturally specific examples, especially from low-resource and dialectal varieties.

Table~\ref{tab:task1-regions} reports performance by language and region. The strongest results appear in the Europe/USA subset, especially English and Italian, which achieve manual evaluation scores of 0.75 and 0.76, respectively. Luxembourgish is a notable exception, with a lower score of 0.43. Classical Arabic and Modern Standard Arabic also perform relatively well, with an average score of 0.62. However, performance drops for Arabic dialects from the Arabian Peninsula, the Levant, and North Africa, with North African dialects being the most challenging among these groups. In the South/East Asian subset, Mandarin performs relatively well, while Hindi and Kashmiri receive the lowest manual evaluation scores, indicating that generated explanations are often unusable. Appendix \ref{app:error-analysis}, Figure \ref{fig:explanation-error} illustrates successful and unsuccessful cases in which models generate correct and incorrect proverb explanations.

\begin{table}[t]
\centering
\tiny
\setlength{\tabcolsep}{4pt}

\begin{threeparttable}

\begin{tabular}{
l
S[table-format=1.4]
S[table-format=1.4]
S[table-format=1.4]
S[table-format=1.4]
}
\toprule
\textbf{Model} &
\textbf{BLEU} &
\textbf{SS} &
\textbf{BERTScore} &
\textbf{ME}\tnote{a} \\
\midrule

Claude Haiku 4.5         & 1.2716 & 0.4540 & 0.8623 & 0.5519 \\
Deepseek V4 Flash        & 1.3797 & \textbf{0.4719} & 0.8666 & 0.6412 \\
\textbf{Gemini 3 Flash}  & 1.0899 & 0.4682 & 0.8688 & \textbf{0.7857} \\
GPT 5.4                  & \textbf{2.1343} & 0.4530 & \textbf{0.8708} & 0.6972 \\
Llama 3.1 Flash 8B       & 0.9886 & 0.3790 & 0.8336 & 0.2524 \\
Mistral V0.2 Instruct 7B & 0.9988 & 0.3923 & 0.8439 & 0.2411 \\
Qwen3 8B                 & 0.9286 & 0.3976 & 0.8591 & 0.3125 \\
\bottomrule

\end{tabular}

\begin{tablenotes}
\footnotesize
\item[a] ME(Manual Evaluation) is computed on a subset $Test_1^{sub} \subset Test_1$; all other metrics are evaluated on the full Test\_1 set.
\end{tablenotes}

\caption{Performance comparison of different models on Task 1 using Test\_1. SS denotes Semantic Textual Similarity, while ME denotes Manual Evaluation score.}
\label{tab:task1-models}

\end{threeparttable}
\end{table}

\subsection{RQ2: Proverb Classification}

\begin{table}[t]
\centering
\scriptsize
\setlength{\tabcolsep}{3.5pt}
\renewcommand{\arraystretch}{1.05}
\begin{tabular}{
@{}p{2.65cm}
S[table-format=1.4]
S[table-format=1.4]
S[table-format=1.4]@{}
}
\toprule
\textbf{Model} &
{\textbf{Acc.}} &
{\textbf{Prec.}} &
{\textbf{F1}} \\
\midrule
Claude Haiku 4.5         
& 0.2874   & 0.3526    & 0.2888 \\
Deepseek V4 Flash        
& 0.3116   & 0.3698    & 0.3217 \\
\textbf{Gemini 3 Flash}  
& \textbf{0.3458}   & \textbf{0.4077}    & \textbf{0.3480} \\
GPT 5.4                  
& 0.3235   & 0.3953    & 0.3250 \\
Llama 3.1 Instruct 8B    
& 0.1552   & 0.3648    & 0.1270 \\
Mistral V0.2 Instruct 7B 
& 0.0974   & 0.3465    & 0.1146 \\
Qwen3 8B                 
& 0.2171   & 0.3360     & 0.2264 \\
\bottomrule
\end{tabular}
\caption{Performance comparison of models on Task~2 using \textbf{\texttt{Test\_1}}.}
\label{tab:task2-models}
\end{table}

Table~\ref{tab:task2-models} shows that semantic classification is difficult for all evaluated models.
Gemini 3 Flash performs best, reaching an F1 score of 0.348, followed by GPT-5.4 with 0.325. These low scores suggest that models struggle to map proverb meanings to higher-level categories such as Ethics and Morality, Friendship, or Social Relations.

Table~\ref{tab:task2-regions} shows similar regional patterns to RQ1. The Europe/USA subset achieves the highest average F1 score, with German obtaining the best language-level result (0.49). Models perform comparatively better on Classical Arabic and Modern Standard Arabic than on regional Arabic dialects. In the South/East Asian subset, Mandarin again performs relatively well, while Hindi and Kashmiri show very low F1 scores.

Overall, these results suggest that proverb classification requires more than recognizing keywords or surface topics. Proverbs are often nuanced, context-dependent, and compatible with multiple overlapping semantic categories, making single-label classification challenging.

\subsection{RQ3: Cross-Cultural Proverb Linking}

\begin{table}[t]
\centering
\tiny
\setlength{\tabcolsep}{4pt}

\begin{threeparttable}

\begin{tabular}{
l
S[table-format=1.4]
S[table-format=1.4]
S[table-format=1.4]
S[table-format=1.4]
}
\toprule
\textbf{Model} &
\textbf{BLEU} &
\textbf{SS} &
\textbf{BertScore} &
\textbf{EM} \\
\midrule

Claude Haiku 4.5         & 2.1570 & 0.8244 & 0.8155 & 0.0085
\\
Deepseek V4 Flash        & 5.9854 & 0.8306 & 0.8234 & 0.0102
 \\
\textbf{Gemini 3 Flash}  & \textbf{7.9817} & \textbf{0.8322} & \textbf{0.8249} & 0.0119
\\
GPT 5.4                  & 5.7451 & 0.8292 & 0.8228 & \textbf{0.0136}
\\
Llama 3.1 Instruct 8B    & 1.3117 & 0.8160 & 0.8125 & 0.0017
\\
Mistral V0.2 Instruct 7B & 0.3083 & 0.7849 & 0.7828 & 0.0000
\\
Qwen3 8B                 & 2.1477 & 0.8190 & 0.8161 & 0.0034
\\
\bottomrule

\end{tabular}
\caption{Performance comparison of different models on Task 3 using Test\_2. SS denotes Semantic Textual Similarity, while EM denotes Exact Match.}
\label{tab:task3-models}
\end{threeparttable}
\end{table}

Table~\ref{tab:task3-models} shows that cross-cultural proverb linking is the most challenging task. Gemini 3 Flash achieves the best overall performance, with a BLEU score of 7.98 and a semantic similarity score of 0.83. GPT-5.4 obtains the highest Exact Match score (0.014), but exact proverb generation remains rare for all models.

Table~\ref{tab:task3-regions} reports performance across languages and regions. The Europe/USA subset is generally the easiest for proverb equivalence generation, with French achieving the highest BLEU score (10.58) and semantic similarity score (0.84). In contrast, South/East Asian languages, especially Hindi and Kashmiri, are among the most difficult target languages. Their low BLEU scores indicate limited success in producing culturally aligned proverb equivalents.

These findings show that models may infer broad pragmatic meaning but often fail to retrieve the conventional proverb used in the target culture. Cross-cultural proverb linking therefore remains a strong test of whether LLMs can move from understanding to culturally appropriate generation. Appendix \ref{app:error-analysis}, Figure \ref{fig:equivalence-error} illustrates examples of successful and unsuccessful cases in which models generated correct and incorrect proverb equivalents.

\section{Conclusion} 

\textsc{WISDOM} provides a multilingual benchmark for evaluating proverb understanding as a form of culturally grounded figurative reasoning. By separating literal wording from pragmatic meaning, assigning shared semantic categories, and linking proverbs through validated English-equivalent proverbs, the dataset supports evaluation beyond surface translation. Our experiments show that current LLMs can often produce plausible explanations for high-resource languages, but they remain unreliable for dialectal, low-resource, and culturally specific proverbs. Classification and cross-cultural proverb linking are especially challenging, suggesting that fluent multilingual generation does not yet imply robust cultural-pragmatic understanding. We hope \textsc{WISDOM} will support future work on multilingual figurative language, culturally aware evaluation, and models that can reason more faithfully across the diverse ways communities encode shared wisdom.

\section*{Limitations}

\textsc{WISDOM} makes several design choices to keep the benchmark controlled, reproducible, and comparable across languages. First, each proverb in \textsc{WISDOMBase} is assigned one primary semantic category. This supports a clear classification task and consistent evaluation, although some proverbs naturally evoke multiple related meanings. To preserve nuance, annotators may provide finer-grained subcategories, and future extensions can support multi-label evaluation.

Second, \textsc{WISDOMBridge} uses validated English-equivalent proverbs as a shared reference point for cross-lingual alignment. This choice enables scalable comparison across many language pairs while avoiding reliance on literal translation. It should not be interpreted as treating English as culturally neutral; rather, English functions as a practical indexing layer. Future versions may add region-specific pivots, such as Modern Standard Arabic for Arabic dialects.

Third, the current benchmark represents proverb equivalence through validated pairwise links. This provides unambiguous test instances for retrieval and generation, but proverb equivalence can also be one-to-many or many-to-many depending on context, register, and usage. Extending \textsc{WISDOMBridge} with graded or clustered equivalence relations is a promising direction.

Fourth, our experiments are limited to relatively small open-source models (7B--8B parameters) due to computational constraints. Consequently, the observed performance gap between proprietary and open-source models is largely driven by a massive discrepancy in parameter scale, rather than the closed-source nature of the systems. The proprietary frontier models we evaluated are estimated to be orders of magnitude larger: for example, GPT-5 class models are estimated at over 1 trillion parameters, and intermediate systems like Gemini 3 Flash and Claude Haiku are estimated in the tens to hundreds of billions. While 70B+ open-weight models are available, they are still significantly smaller than the closed frontier systems, and evaluating them was not feasible under our resource constraints. We explicitly highlight this scale difference to prevent misinterpreting the results as a simple comparison of open vs. closed weights. Evaluating larger open-weight systems as they become more computationally accessible will help isolate the effects of scale from access constraints.

Fifth, we evaluate models in a monolingual zero-shot setting to measure their basic ability to understand proverbs without additional fine-tuning. However, this setting may not fully reflect performance on complex figurative language tasks. Future work could improve results by fine-tuning models on WisdomBase and WisdomBridge and exploring alternative prompting strategies, such as chain-of-thought prompting, original-language prompting, and few-shot learning.

Finally, automatic metrics only approximate cultural and pragmatic adequacy. We therefore report multiple complementary metrics and include human evaluation for proverb explanation. The results should be interpreted as comparative evidence of model behavior rather than as a complete measure of cultural understanding.

\section*{Use of AI Assistance}
In preparing this work, we used AI-assisted tools for code completion with GitHub Copilot and language editing, such as spell checking and stylistic revisions with Grammarly and ChatGPT. The authors take full responsibility for the scientific content, analyses, and conclusions presented in this work.

\bibliography{custom}

\appendix
\clearpage

\section*{Appendices}
\addcontentsline{toc}{section}{Appendices}

The appendices provide supporting documentation and results:
\begin{itemize}
    
    \item Appendix~\ref{app:dataset-protocol}: dataset collection, annotation protocol, and schema.
    \item Appendix~\ref{app:related_work}: comparison with related work.
    \item Appendix~\ref{app:statistics-alignment}: dataset statistics and cross-lingual alignment coverage.
    \item Appendix~\ref{app:manual-eval}: manual evaluation protocol.
    \item Appendix~\ref{app:implementation-details}: experimental implementation details.
    \item Appendix~\ref{app:prompts}: evaluation prompts.
    \item Appendix~\ref{app:additional-results}: detailed language and region results.
    \item Appendix~\ref{app:examples}: selected examples from \textsc{WISDOMBridge}.
    \item Appendix~\ref{app:error-analysis}: cases of success and failure in proverb explanation and equivalence.
\end{itemize}

\section{Dataset Collection and Annotation Protocol}
\label{app:dataset-protocol}

\subsection{Collection Sources by Region}
\label{app:collection-protocol}

This appendix summarizes the language-specific collection process used to construct \textsc{WISDOM}. Because proverb traditions vary in digitization, regional variation, written documentation, and reliance on oral transmission, contributors documented source types, cleaning steps, and validation procedures for each language or dialectal variety. Across all subsets, annotators followed the shared \textsc{WISDOM} protocol: preserve the original proverb form, separate literal wording from pragmatic meaning, provide original-language and English explanations, assign semantic categories, and link proverbs to English equivalents when a true meaning-based correspondence was available.

\paragraph{Arabic Varieties.}
Arabic is the largest and most internally diverse part of \textsc{WISDOM}. The dataset includes Classical Arabic, Modern Standard Arabic (MSA), and regional varieties from the Arabian Peninsula, the Levant, and North Africa. The Classical Arabic subset contains 132 proverbs, while the MSA subset contains 620 proverbs collected from classical Arabic proverb collections and reviewed for explanation quality. The Saudi subset contains 1,161 proverbs collected mainly from digital cultural and heritage resources covering social practices, folklore, and regional traditions. The Kuwaiti, Qatari, Bahraini, and general Arabian Peninsula subsets were collected from printed compilations, regional anthologies, online proverb portals, cultural magazines, and historical sources; contributors removed duplicates, malformed entries, and items with unreliable dialect attribution.

The Emirati subset contains 305 proverbs collected from UAE proverb and heritage materials, including regional proverb compilations and publicly available cultural resources. The Omani subset contains 103 proverbs gathered from Omani proverb collections and public cultural resources. For both subsets, native speakers checked dialectal wording, removed duplicates or incomplete entries, verified English explanations, and validated semantic categories and English-equivalent links.

The Levantine portion includes Syrian, Jordanian, and Palestinian Arabic. The Syrian subset contains 1,148 proverbs drawn from a printed collection of Suwayda-region folk expressions, preserving dialectal forms and Arabic explanatory glosses. The Jordanian subset contains 1,141 proverb--explanation pairs extracted from a public source using OCR, while the Palestinian subset contains 5,650 pairs collected from a Hindawi resource. For Jordanian and Palestinian Arabic, Arabic explanations were translated into English, and model-assisted topic suggestions were manually checked by contributors.

The North African portion includes Algerian, Egyptian, Libyan, Mauritanian, Moroccan, and Tunisian Arabic. The Moroccan subset contains 2,290 proverbs derived from \textit{Popular Moroccan Proverbs} (\ar{الأمثال الشعبية المغربية}) using OCR and structured parsing. Algerian proverbs were drawn from \ar{الأمثال الشعبية الجزائرية}, contributing 1,003 proverbs with explanations and original thematic categories. The Tunisian subset contains 1,160 proverbs from thematically organized Tunisian proverb materials, and the Libyan subset contains 1,284 proverbs from Libyan proverb collections. The Mauritanian subset contains 532 proverbs selected from a governmental Mauritanian cultural resource. Cleaning involved normalizing inconsistent metadata, reviewing incomplete annotations, consolidating duplicate or near-duplicate variants, and preserving the original dialectal forms.

\paragraph{European and USA Subsets.}
The European/USA portion includes English, French, Luxembourgish, German, Portuguese, Spanish, Catalan, Italian, and Danish. English contains 1,988 proverbs and serves as the main equivalent inventory for \textsc{WISDOMBridge}. English entries were curated from public-domain proverb collections, proverb dictionaries, and online paremiological resources; duplicate spelling variants were consolidated when they represented the same conventional proverb.

French, Luxembourgish, German, Portuguese, Italian, and Danish were collected from language-specific proverb dictionaries, public-domain collections, educational or cultural proverb resources, and manually curated online paremiological lists. Native speakers reviewed the original proverb forms, removed malformed or duplicated entries, produced or corrected English explanations, and validated category assignments. For \textsc{WISDOMBridge}, contributors linked each proverb to an English proverb only when the English item expressed the same pragmatic lesson, rejecting links based only on literal overlap or topical similarity.

Spanish contains 2,501 proverbs collected from Spanish proverb collections and multilingual paremiological resources. Catalan contains 377 proverbs and follows the same validation process. For both Spanish and Catalan, candidate English equivalents were proposed automatically and then manually revised to keep only reliable meaning-based correspondences.

\paragraph{South and East Asian Subsets.}
The South and East Asian portion includes Mandarin Chinese, Hindi, and Kashmiri. Mandarin Chinese contains 1,152 proverbs focused on social, military, and political themes, extracted from a published collection of Chinese proverbs. This subset includes proverb traditions such as \textit{yanyu} and \textit{chengyu}, which are often rooted in classical literature, historical chronicles, and philosophical traditions. Hindi contains 747 proverbs reflecting proverb traditions shaped by literature, education, religion, cinema, and popular culture. Kashmiri contains 747 proverbs and represents an underrepresented South Asian language with a proverb tradition connected to poetry, folklore, music, and Kashmiriyat. Native speakers reviewed meaning explanations and category assignments for these subsets.

\subsection{\textsc{WISDOMBase} Schema}
\label{app:wisdom-schema}

\textsc{WISDOMBase} stores proverb-level annotations used for explanation and classification. Table~\ref{tab:wisdom-schema} lists the core fields. The schema separates original proverb wording from meaning explanations so that evaluation targets pragmatic interpretation rather than surface translation. Table~\ref{tab:wisdom-categories} gives the shared macro-category inventory used for the classification task.

\begin{table*}[t]
\centering
\small
\begin{tabular}{p{0.22\linewidth}p{0.70\linewidth}}
\hline
\textbf{Field} & \textbf{Description} \\
\hline
\texttt{ID} & Unique alphanumeric identifier for each proverb. \\
\texttt{language} & Macro-language of the proverb. For Arabic and other regionally variable languages, annotators record the dialect or regional variety when applicable. \\
\texttt{proverb} & Original proverb, transcribed as it appears in the source or oral tradition. Dialectal spelling and local orthography are preserved. \\
\texttt{explanation\_original} & Pragmatic explanation in the original language, or in the standard variety where appropriate. For dialectal Arabic, annotators prioritize an MSA explanation to support monolingual semantic retrieval. \\
\texttt{explanation\_en} & English explanation of the pragmatic meaning in one or two sentences, focusing on the lesson or general human situation rather than the literal imagery. \\
\texttt{category} & Primary semantic macro-category selected from the shared inventory. \\
\texttt{subcategory} & Optional fine-grained label, such as marriage, hypocrisy, haste, friendship, or false pride. \\
\texttt{source} & Source information, such as a book title, URL, or oral tradition notes when available. \\
\texttt{validated\_by} & Native-speaker annotator or validator identifier, used for internal quality tracking. \\
\hline
\end{tabular}
\caption{Core fields in \textsc{WISDOMBase}.}
\label{tab:wisdom-schema}
\end{table*}

\begin{table*}[t]
\centering
\small
\begin{tabular}{p{0.05\linewidth}p{0.39\linewidth}p{0.05\linewidth}p{0.39\linewidth}}
\hline
\textbf{No.} & \textbf{Macro-category} & \textbf{No.} & \textbf{Macro-category} \\
\hline
1 & Fate, Religion \& Providence & 11 & Time, Change \& Life Circumstances \\
2 & Wisdom, Knowledge \& Experience & 12 & Nature, Animals \& Environment \\
3 & Character, Ethics \& Morality & 13 & Daily Life, Health \& Material Living \\
4 & Human Nature, Emotions \& Psychological States & 14 & Success, Strategy \& Ambition \\
5 & Family, Love \& Gender & 15 & Humor, Irony, Paradox \& Folly \\
6 & Friendship, Conflict \& Social Relations & 16 & Life, Death \& Existential Reflection \\
7 & Social Norms, Etiquette \& Customs & 17 & Hospitality, Generosity \& Reciprocity \\
8 & Communication \& Speech & 18 & Luck \& Chance \\
9 & Power, Justice \& Governance & 19 & Appearance vs. Reality \\
10 & Wealth, Work \& Economic Life & 20 & Self \& Identity \\
\hline
\end{tabular}
\caption{Semantic macro-categories used for \textsc{WISDOMBase} classification.}
\label{tab:wisdom-categories}
\end{table*}

\subsection{\textsc{WISDOMBridge} Alignment Fields}
\label{app:bridge-schema}

\textsc{WISDOMBridge} represents cross-cultural proverb equivalence through validated links to English-equivalent proverbs. These links are used to derive source--target retrieval pairs for Task~3. Table~\ref{tab:bridge-schema} summarizes the fields used when constructing these pairs.

\begin{table*}[t]
\centering
\small
\begin{tabular}{p{0.22\linewidth}p{0.70\linewidth}}
\hline
\textbf{Field} & \textbf{Description} \\
\hline
\texttt{source\_language} & Language or regional variety of the source proverb. \\
\texttt{source\_proverb} & Original source-language proverb. \\
\texttt{english\_equivalent} & Validated English proverb that expresses the same pragmatic lesson as the source proverb. \\
\texttt{target\_language} & Language or regional variety used as the target in a retrieval pair. \\
\texttt{target\_proverb} & Target-language proverb linked to the same English-equivalent proverb. \\
\texttt{validation\_status} & Internal validation marker indicating whether the pair was manually accepted, rejected, or flagged for review. \\
\hline
\end{tabular}
\caption{Core fields used to construct source--target pairs in \textsc{WISDOMBridge}.}
\label{tab:bridge-schema}
\end{table*}

\section{Comparison with Existing Proverb Datasets}
\label{app:related_work}

While prior datasets have advanced proverb understanding through tasks such as explanation, metaphor analysis, classification, translation, and narrative generation, they typically focus on a single language or a limited set of language pairs, and few provide explicit semantic categorization and systematic cross-lingual equivalence annotations. \textsc{WISDOM} addresses these limitations by bringing together 39,838 instances across 12 languages and 18 varieties, organized into 20 semantic categories, with equivalence links spanning all covered languages and varieties. Furthermore, it supports a unified evaluation framework covering proverb explanation, semantic classification, and cross-lingual equivalence. Table \ref{tab:related-work} provides a detailed comparison of \textsc{WISDOM} with existing proverb datasets.

\begin{table*}[t]
\centering
\scriptsize
\setlength{\tabcolsep}{4pt}
\renewcommand{\arraystretch}{1.15}

\begin{tabularx}{\textwidth}{
    l
    X
    c
    X
    X
    X
}
\toprule
\textbf{Dataset} &
\textbf{Languages} &
\textbf{Size} &
\textbf{Categories} &
\textbf{Equivalence} &
\textbf{Tasks} \\
\midrule
Jawaher (Magdy et al., 2025)
& Arabic and its dialects
& 10,035 (2,500 explanations)
& --
& Arabic--English
& Translation and explanation \\

Promotheus (Gozde et al., 2016)
& English and Italian
& 1,054
& --
& English--Italian
& Explanation and metaphor analysis \\

Prone (Thapa et al., 2025)
& Nepali
& 2,830
& 5 categories
& --
& Explanation and classification \\

ePiC (Sayan et al., 2022)
& English
& 2,500
& --
& --
& Proverb recommendation and narrative generation \\

BRoverbs (Thalis et al., 2025)
& Portuguese
& 579 (3 stories per proverb)
& --
& --
& Proverb-to-story and story-to-proverbs \\

ProverbIT (Mensa et al., 2025)
& Italian
& 100
& --
& --
& Proverb completion \\

\textbf{WISDOM (ours)}
& \textbf{30 languages and varieties}
& \textbf{39,838}
& \textbf{20 categories}
& \textbf{30 languages and varieties}
& \textbf{Explanation, classification, and equivalence} \\
\bottomrule
\end{tabularx}
\caption{Comparison of existing proverb datasets with WISDOM.}
\label{tab:related-work}
\end{table*}

\section{Dataset Statistics and Cross-Lingual Alignment Coverage}
\label{app:statistics-alignment}

\textsc{WISDOM} is intentionally multilingual and regionally diverse rather than fully balanced. Table~\ref{tab:wisdom-statistics} summarizes the overall coverage. These statistics contextualize the evaluation results: some languages have larger written proverb traditions and more digitized sources, whereas others are represented by smaller curated collections.

\begin{table*}[t]
\centering
\small
\begin{tabular}{p{0.24\linewidth}p{0.68\linewidth}}
\hline
\textbf{Statistic} & \textbf{Value and interpretation} \\
\hline
Total size & 39,838 proverb entries. \\
Coverage & 30 language and regional varieties, grouped by macro-language, region, and cultural area. \\
Regional coverage & Classical Arabic and MSA; Arabic varieties from the Arabian Peninsula, the Levant, and North Africa; European/USA subsets including French, Luxembourgish, German, Portuguese, Spanish, Catalan, Italian, Danish, and English; and South/East Asian subsets including Mandarin, Hindi, and Kashmiri. \\
Largest subsets & Palestinian Arabic (5,650), Yemeni Arabic (3,506), Spanish (2,501), Moroccan Arabic (2,290), and English (1,988). \\
Long-tail subsets & Omani Arabic (103), Luxembourgish (303), Emirati Arabic (305), Catalan (377), Bahraini Arabic (530), and Mauritanian Arabic (532). \\
Annotation scope & Each proverb is annotated with its original form, pragmatic explanation, English explanation, semantic category, and, where available, cross-lingual equivalence links. \\
\hline
\end{tabular}
\caption{Dataset statistics and coverage for \textsc{WISDOM}.}
\label{tab:wisdom-statistics}
\end{table*}

\section{Quality Control and Annotator Agreement}\label{app:quality} 
Quality control combined native-speaker review with automated consistency checks. The benchmark was constructed and validated by a distributed team of 14 native-speaking expert annotators, each assigned to their native language or regional variety to ensure cultural authenticity. 
Contributors first verified that each proverb was a conventional expression in the relevant language or dialect and that the original form preserved local orthography and dialectal features. They then checked that the English explanation described the pragmatic lesson rather than a literal translation. Category assignments were reviewed against the shared 20-category inventory, and uncertain cases were discussed during coordination meetings to establish consensus. For \textsc{WISDOMBridge}, annotators validated direct links between non-English proverbs and English-equivalent proverbs. A link was accepted only when both proverbs expressed the same lesson in normal use. Literal translations, broad topical matches, and weak paraphrases were rejected. The final processing scripts checked for missing required fields, duplicate identifiers, invalid category names, empty explanations, and inconsistent English-equivalent links.
\paragraph{Inter-Annotator Agreement (IAA).}

Due to the cultural and dialectal specialization of the benchmark, each proverb and its translation were initially annotated and validated by a single native-speaker expert assigned to the corresponding language variety. Inter-annotator agreement (IAA) was subsequently assessed on nine intersection languages for which annotators were able to validate samples from language varieties they had not annotated during the initial annotation round. These were the only languages for which overlapping annotations were available. For each language, IAA was evaluated on 100 samples for two tasks: identifying equivalent proverbs and classifying proverbs into categories. Results are displayed in Tables \ref{tab:iaa_category} \& \ref{tab:iaa-equivalent}.

For equivalent proverb identification, Cohen's $\kappa$ ranged from 0.582 to 0.970, with agreement rates between 59\% and 97\% across the eight languages. The highest agreement was observed for France ($\kappa = 0.970$, 97\%), followed by Bahrain ($\kappa = 0.868$, 87\%) and Qatar ($\kappa = 0.858$, 86\%). The lowest agreement was observed for China ($\kappa = 0.582$, 59\%).

For proverb category classification, Cohen's $\kappa$ ranged from 0.467 to 0.725, with agreement rates between 50\% and 75\%. Tunisia achieved the highest agreement ($\kappa = 0.725$, 75\%), followed by France ($\kappa = 0.537$, 59\%) and Mandarin ($\kappa = 0.501$, 57\%). The lowest agreement was observed for Syria ($\kappa = 0.467$, 50\%), closely followed by Italy ($\kappa = 0.467$, 50\%).

Overall, the results indicate higher inter-annotator agreement for equivalent proverb identification than for category classification. This difference suggests that annotators were generally more consistent in determining whether two proverbs were equivalent than in assigning proverbs to semantic categories, where interpretations and category boundaries may be more subjective. Moreover, a proverb may belong to multiple categories, introducing additional ambiguity into the classification task and potentially contributing to the lower agreement observed.

\begin{table}[t]
\centering
\scriptsize
\begin{tabular}{lrrr}
\toprule
Language & Examples & Cohen's $\kappa$ & Agreement (\%) \\
\midrule
Tunisia   & 100 & 0.725 & 75 \\
Egypt     & 100 & 0.495 & 53 \\
France    & 100 & 0.537 & 59 \\
Bahrain   & 100 & 0.492 & 53 \\
Italy     & 100 & 0.467 & 50 \\
Qatar     & 100 & 0.501 & 54 \\
Syria     & 100 & 0.467 & 50 \\
Mandarin  & 100 & 0.501 & 57 \\
Palestine & 100 & 0.477 & 53 \\
\midrule
AVG     & 100 & 0.518 & 56   \\
\bottomrule
\end{tabular}
\caption{Inter-annotator agreement in proverb category classification. For each language, agreement was assessed on 100 samples.}
\label{tab:iaa_category}
\end{table}

\begin{table}[t]
\centering
\scriptsize
\begin{tabular}{lrrr}
\toprule
Language & Examples & Cohen's $\kappa$ & Agreement (\%) \\
\midrule
China   & 100 & 0.582 & 59 \\
Tunisia & 100 & 0.818 & 82 \\
Egypt   & 100 & 0.597 & 60 \\
France  & 100 & 0.970 & 97 \\
Bahrain & 100 & 0.868 & 87 \\
Italy   & 100 & 0.718 & 72 \\
Qatar   & 100 & 0.858 & 86 \\
Syria   & 100 & 0.637 & 64 \\
Palestine & 100 & 0.806 & 81 \\

\midrule
Avg   & 100 & 0.762 & 76.44   \\
\bottomrule
\end{tabular}
\caption{Inter-annotator agreement in finding equivalent proverbs. For each language, agreement was assessed on 100 samples.}
\label{tab:iaa-equivalent}
\end{table}

\section{Manual Evaluation Protocol}
\label{app:manual-eval}
\label{sec:manual-evaluation}

Manual evaluation was conducted for the proverb explanation task to complement automatic similarity metrics. Each reviewer evaluated proverbs from their native language or dialect. For each item, the reviewer saw the original proverb, the gold English explanation, and anonymized explanations generated by the evaluated models. Reviewers judged whether the model output captured the intended cultural use of the proverb rather than merely translating its words.

\paragraph{Scoring Task.}
Reviewers assigned each model output a score from 1 to 5 according to the following scale:
\begin{itemize}
    \item \textbf{5 -- Flawless:} Fully correct cultural meaning, concise, no filler, and no repetition of the proverb.
    \item \textbf{4 -- Minor issue:} Mostly correct but slightly lacking in depth or containing minor formatting issues.
    \item \textbf{3 -- Good but flawed:} Basic meaning is correct, but the explanation lacks cultural depth or includes filler or repetition.
    \item \textbf{2 -- Major violations:} Incorrect cultural context, overly generic explanation, literal translation, or unnecessary repetition of the proverb.
    \item \textbf{1 -- Unusable:} Completely incorrect, hallucinated, wrong language, or irrelevant.
\end{itemize}

Reviewers were instructed to penalize invented cultural facts, word-for-word translations, generic Western replacements that were not culturally accurate, and filler phrases such as ``This proverb means''. The central criterion was whether the explanation clearly showed when and why a native speaker would use the proverb in real life. Scores were normalized to $[0,1]$ for reporting alongside the automatic metrics.

\section{Implementation Details}\label{app:implementation-details}

We ran the experiments described in Section~\ref{sec:experiments} on a workstation equipped with an Intel(R) Core(TM) Ultra 9 285K processor (3.70 GHz), 128 GB RAM, and an NVIDIA GeForce RTX 5090 GPU with 32 GB of memory. Experiments were implemented using the Hugging Face ecosystem\footnote{https://huggingface.co/}, including \textit{Transformers} for model inference, \textit{Datasets} for dataset handling, and \textit{Evaluate} for metric computation. For open-weights models, inference utilized 4-bit NormalFloat (`nf4`) quantization with `bfloat16` compute dtype via `bitsandbytes`, while decoding hyperparameters such as temperature and top-$p$ relied on the respective provider and model defaults; The inference and evaluation codebase is available in our GitHub repository\footnote{Code is available at
\url{https://github.com/salmane11/WISDOM}}.

\section{Evaluation Prompts}
\label{app:prompts}
\label{sec:prompts}

This appendix lists the prompts used for proverb explanation, proverb classification, and proverb linking. In the explanation task, the model receives a proverb and produces a concise English explanation of its figurative meaning as used by native speakers. In the classification task, the model receives a proverb and the 20-category inventory, then selects the most appropriate category. In the linking task, the model receives a source proverb and target language, then generates the target-language proverb that conveys the closest equivalent meaning. Across all tasks, prompts emphasize cultural grounding and discourage literal translation.

\begin{figure*}
\begin{lstlisting}[basicstyle=\small\ttfamily, captionpos=b, backgroundcolor=\color{gray!10},numbers=left, language=Python, tabsize=1, xleftmargin=0.5em, caption={Proverbs Explanation Prompt}, label={lst:task1_prompt}]

You are an expert in world proverbs, cultural anthropology, and linguistics.

Task: Provide a concise cultural explanation of the following proverb.

Context:
- Language/Culture: "{language}"
- Proverb: "{proverb}"

Requirements:
1. Output Format: Provide EXACTLY one informative sentence. Do not include introductory text, quotes, or conversational filler. Do not repeat the proverb in the Explanation.
2. Content: Explain the core meaning and how it is applied within its native culture. 
3. Accuracy & Constraints: Write entirely in English. Be strictly accurate to the source culture. Do not invent historical or cultural context.

Explanation: 

\end{lstlisting}
\end{figure*}
\begin{figure*}
\begin{lstlisting}[basicstyle=\small\ttfamily, captionpos=b, backgroundcolor=\color{gray!10},numbers=left, language=Python, tabsize=1, xleftmargin=0.5em, caption={Proverbs Classification Prompt}, label={lst:task2_prompt}]

You are an expert in world proverbs, cultural anthropology, and linguistics.

Task: Select exactly one category from the list below that best describes the given proverb.

Context:
- Language/Culture: "{language}"
- Proverb: "{proverb}"

Possible categories:
- Fate, Religion & Providence
- Wisdom, Knowledge & Experience
- Character, Ethics & Morality
- Human Nature, Emotions & Psychological States
- Family, Love & Gender
- Friendship, Conflict & Social Relations
- Social Norms, Etiquette & Customs
- Communication & Speech
- Power, Justice & Governance
- Wealth, Work & Economic Life
- Time, Change & Life Circumstances
- Nature, Animals & Environment
- Daily Life, Health & Material Living
- Success, Strategy & Ambition
- Humor, Irony, Paradox & Folly
- Life, Death & Existential Reflection
- Hospitality, Generosity & Reciprocity
- Luck & Chance
- Appearance vs. Reality
- Self & Identity

Requirements:
1. Output Format: Provide EXACTLY the name of the chosen category. Do not include introductory text, conversational filler, punctuation, or any additional information.
2. Accuracy & Constraints: Write the category exactly as it is written in the provided list.

Category:

\end{lstlisting}
\end{figure*}
\begin{figure*}
\begin{lstlisting}[basicstyle=\small\ttfamily, captionpos=b, backgroundcolor=\color{gray!10},numbers=left, language=Python, tabsize=1, xleftmargin=0.5em, caption={Proverbs Linking Prompt}, label={lst:task3_prompt}]

You are an expert in world proverbs, comparative literature, and cultural linguistics.

Task: Identify the closest culturally equivalent proverb in the target language that conveys the exact same core meaning, moral, or lesson as the source proverb.

Context:
- Source Language: "{source_language}"
- Proverb: "{proverb}"
- Target Language: "{target_language}"

Requirements:
1. Output Format: Provide ONLY the text of the equivalent proverb in the target language. Do not include introductory text, explanations, literal translations, quotation marks, or conversational filler.
2. Constraint: Do not invent or literally translate the source proverb to force a match; it must be a recognized, authentic saying in the target language culture.

Equivalent Proverb: 


\end{lstlisting}
\end{figure*}

\section{Additional Experimental Results}
\label{app:additional-results}

The following tables report language- and region-level results for the three tasks. These tables complement the model-level results in the main paper and make the regional performance patterns explicit.

\paragraph{Proverb Explanation.}
Table~\ref{tab:task1-regions} reports explanation performance by language and region. The table shows where models produce culturally adequate explanations and where performance drops for dialectal or lower-resource varieties.

\begin{table*}[t]
\centering
\small
\setlength{\tabcolsep}{5pt}
\begin{threeparttable}

\begin{tabular}{
l
l
S[table-format=1.4]
S[table-format=1.4]
S[table-format=1.4]
S[table-format=1.4]
}
\toprule
\textbf{Region} &
\textbf{Language} &
\textbf{BLEU} &
\textbf{SS} &
\textbf{BERTScore} &
\textbf{Manual Eval.}\tnote{a} \\
\midrule

\multirow{2}{*}{Arab World}
& Classical Arabic        & 1.9190 & 0.5159 & 0.8540 & 0.6179 \\
& Modern Standard Arabic  & 1.7396 & 0.4738 & 0.8636 & 0.6321 \\
\cmidrule(lr){1-6}
\textit{Average}
&                         & 1.8293 & 0.4948 & 0.8588 & 0.6250 \\
\midrule

\multirow{8}{*}{Arabian Peninsula}
& Peninsular Arabic       & 0.7359 & 0.3918 & 0.8483 & 0.4036 \\
& Bahraini Arabic         & 1.1398 & 0.4265 & 0.8437 & 0.3893 \\
& Kuwaiti Arabic          & 1.3324 & 0.4338 & 0.8514 & 0.4214 \\
& Omani Arabic            & 1.1511 & 0.4236 & 0.8650 & 0.5392 \\
& Qatari Arabic           & 0.6435 & 0.4329 & 0.8415 & 0.4785 \\
& Saudi Arabic            & 0.6446 & 0.4537 & 0.8604 & 0.5142 \\
& Emirati Arabic          & 0.8128 & 0.4117 & 0.8536 & 0.2714 \\
& Yemeni Arabic           & 0.7850 & 0.3849 & 0.8550 & 0.3304 \\
\cmidrule(lr){1-6}
\textit{Average}
&                         & 0.9056 & 0.4199 & 0.8524 & 0.4208 \\
\midrule

\multirow{3}{*}{Levant}
& Jordanian Arabic        & 0.9361 & 0.5257 & 0.8542 & 0.4999 \\
& Palestinian Arabic      & 0.6582 & 0.3654 & 0.8556 & 0.3821 \\
& Syrian Arabic           & 0.5936 & 0.3528 & 0.8529 & 0.4857 \\
\cmidrule(lr){1-6}
\textit{Average}
&                         & 0.7293 & 0.4146 & 0.8542 & 0.4559 \\
\midrule

\multirow{6}{*}{North Africa}
& Algerian Arabic         & 0.6065 & 0.3851 & 0.8504 & 0.3750 \\
& Egyptian Arabic         & 1.2183 & 0.4548 & 0.8541 & 0.6536 \\
& Libyan Arabic           & 0.9915 & 0.4127 & 0.8593 & 0.3786 \\
& Mauritanian Arabic      & 0.7537 & 0.3337 & 0.8522 & 0.2249 \\
& Moroccan Arabic         & 1.0852 & 0.4512 & 0.8596 & 0.3428 \\
& Tunisian Arabic         & 0.8711 & 0.4265 & 0.8574 & 0.5107 \\
\cmidrule(lr){1-6}
\textit{Average}
&                         & 0.9210 & 0.4107 & 0.8555 & 0.4143 \\
\midrule

\multirow{9}{*}{Europe and USA}
& Catalan                 & 0.6406 & 0.3644 & 0.8663 & 0.5786 \\
& Danish                  & 2.7706 & 0.5636 & 0.8714 & 0.6393 \\
& English                 & 1.7662 & 0.5534 & 0.8756 & 0.7500 \\
& French                  & 2.0119 & 0.5448 & 0.8696 & 0.7464 \\
& German                  & 2.7564 & 0.6406 & 0.8775 & 0.7000 \\
& Italian                 & 2.4820 & 0.5830 & 0.8726 & 0.7678 \\
& Luxembourgish           & 0.4344 & 0.2942 & 0.8571 & 0.4286 \\
& Portuguese              & 2.0746 & 0.5644 & 0.8728 & 0.7357 \\
& Spanish                 & 1.0354 & 0.4387 & 0.8698 & 0.6500 \\
\cmidrule(lr){1-6}
\textit{Average}
&                         & \textbf{1.7747} & \textbf{0.5052} & \textbf{0.8703} & \textbf{0.6662} \\
\midrule

\multirow{3}{*}{South Asia}
& Hindi                   & 0.1410 & 0.1708 & 0.8307 & 0.1678 \\
& Kashmiri                & 0.1355 & 0.1645 & 0.8356 & 0.1035 \\
& Mandarin                & 1.1754 & 0.4193 & 0.8646 & 0.6678 \\
\cmidrule(lr){1-6}
\textit{Average}
&                         & 0.4840 & 0.2515 & 0.8436 & 0.3130 \\
\bottomrule
\end{tabular}

\begin{tablenotes}
\footnotesize
\item[a] Manual evaluation is computed on a subset
$Test_1^{sub} \subset Test_1$; all other metrics are evaluated on the full $Test_1$ set.
\end{tablenotes}
\caption{Mean explanation performance grouped by language and region, measured using BLEU, Semantic Similarity (SS), BERTScore, and manual evaluation.}
\label{tab:task1-regions}

\end{threeparttable}
\end{table*}

\paragraph{Proverb Classification.}
Table~\ref{tab:task2-regions} reports classification performance by language and region. These results help identify whether models can abstract from proverb wording to the intended semantic category across different cultural areas.

\begin{table*}[t]
\centering
\small
\setlength{\tabcolsep}{5pt}

\begin{threeparttable}

\begin{tabular}{
l
l
S[table-format=1.4]
S[table-format=1.4]
S[table-format=1.4]
}
\toprule
\textbf{Region} &
\textbf{Language} &
\textbf{Accuracy} &
\textbf{Precision} &
\textbf{F1} \\
\midrule

\multirow{2}{*}{Arab World}
& Classical Arabic         & 0.3743   & 0.4014    & 0.3580 \\
& Modern Standard Arabic   & 0.2514   & 0.3701    & 0.2732 \\
\cmidrule(lr){1-5}
\textit{Average}
&                           & 0.3129   & 0.3857    & 0.3156 \\
\midrule

\multirow{8}{*}{Arabian Peninsula}
& Peninsular Arabic       & 0.1629   & 0.3362    & 0.1519 \\
& Bahraini Arabic         & 0.2214   & 0.2815    & 0.2142 \\
& Kuwaiti Arabic          & 0.3214   & 0.4613    & 0.3174 \\
& Omani Arabic            & 0.2300   & 0.4198    & 0.2207 \\
& Qatari Arabic           & 0.2657   & 0.3601    & 0.2652 \\
& Saudi Arabic            & 0.1671   & 0.2732    & 0.1812 \\
& Emirati Arabic          & 0.1214   & 0.3341    & 0.1387 \\
& Yemeni Arabic           & 0.1429   & 0.3026    & 0.1648 \\
\cmidrule(lr){1-5}
\textit{Average}
&                           & 0.2041   & 0.3461    & 0.2067 \\
\midrule

\multirow{3}{*}{Levant}
& Jordanian Arabic        & 0.2457   & 0.4771    & 0.2772 \\
& Palestinian Arabic      & 0.1457   & 0.3273    & 0.1671 \\
& Syrian Arabic           & 0.1557   & 0.2463    & 0.1516 \\
\cmidrule(lr){1-5}
\textit{Average}
&                           & 0.1824   & 0.3502    & 0.1987 \\
\midrule

\multirow{6}{*}{North Africa}
& Algerian Arabic         & 0.1671   & 0.3365    & 0.1923 \\
& Egyptian Arabic         & 0.3071   & 0.4065    & 0.3053 \\
& Libyan Arabic           & 0.1357   & 0.2994    & 0.1288 \\
& Mauritanian Arabic      & 0.1443   & 0.1882    & 0.1384 \\
& Moroccan Arabic         & 0.1157   & 0.1721    & 0.1176 \\
& Tunisian Arabic         & 0.3629   & 0.4714    & 0.3630 \\
\cmidrule(lr){1-5}
\textit{Average}
&                           & 0.2055   & 0.3123    & 0.2076 \\
\midrule

\multirow{9}{*}{Europe and USA}
& Catalan                 & 0.1671   & 0.2546    & 0.1846 \\
& Danish                  & 0.4029   & 0.4800    & 0.4000 \\
& English                 & 0.3629   & 0.5267    & 0.3602 \\
& French                  & 0.3357   & 0.5712    & 0.3402 \\
& German                  & \textbf{0.4743} & \textbf{0.6417} & \textbf{0.4861} \\
& Italian                 & 0.45     & 0.5185    & 0.4505 \\
& Luxembourgish           & 0.2729   & 0.3754    & 0.2821 \\
& Portuguese              & 0.4029   & 0.4952    & 0.3845 \\
& Spanish                 & 0.2543   & 0.3541    & 0.2662 \\
\cmidrule(lr){1-5}
\textit{Average}
&                           & \textbf{0.347}    & \textbf{0.4686}    & \textbf{0.3505} \\
\midrule

\multirow{3}{*}{South Asia}
& Hindi                   & 0.0643   & 0.0773    & 0.0601 \\
& Kashmiri                & 0.0700   & 0.1139    & 0.0636 \\
& Mandarin                & 0.4014   & 0.5372    & 0.4143 \\
\cmidrule(lr){1-5}
\textit{Average}
&                           & 0.1786   & 0.2428    & 0.1793 \\
\bottomrule

\end{tabular}
\caption{Mean classification performance grouped by language and region, measured using accuracy, precision, and weighted F1-score.}
\label{tab:task2-regions}
\end{threeparttable}
\end{table*}

\paragraph{Cross-Cultural Proverb Linking.}
Table~\ref{tab:task3-regions} reports linking performance by language and region. This analysis indicates which target languages are easier or harder for models when generating culturally conventional equivalent proverbs.

\begin{table*}[t]
\centering
\small
\setlength{\tabcolsep}{5pt}

\begin{threeparttable}

\begin{tabular}{
l
l
S[table-format=2.4]
S[table-format=1.4]
S[table-format=1.4]
S[table-format=1.4]
}
\toprule
\textbf{Region} &
\textbf{Language} &
\textbf{BLEU} &
\textbf{SS} &
\textbf{BERTScore} &
\textbf{EM} \\
\midrule

\multirow{2}{*}{Arab World}
& Classical Arabic         & 1.6521 & 0.8095 & 0.7965 & 0.0000
 \\
& Modern Standard Arabic   & 1.9083 & 0.8156 & 0.8156 & 0.0000
\\
\cmidrule(lr){1-6}
\textit{Average}
&                           & 1.7802 & 0.8126 & 0.8061 & 0.0000
 \\
\midrule

\multirow{7}{*}{Arabian Peninsula}
& Bahraini Arabic         & 1.2268 & 0.8085 & 0.8009 & 0.0000 \\
& Kuwaiti Arabic          & 8.0355 & 0.8334 & 0.8170 & 0.0357 \\
& Omani Arabic            & 0.9467 & 0.8072 & 0.8014 & 0.0000 \\
& Qatari Arabic           & 7.2741 & 0.8209 & 0.8142 & 0.0071 \\
& Saudi Arabic            & 0.5643 & 0.8040 & 0.7981 & 0.0000 \\
& Emirati Arabic          & 1.1131 & 0.8167 & 0.8192 & 0.0000 \\
& Yemeni Arabic           & 5.9545 & 0.8283 & 0.8205 & 0.0071 \\
\cmidrule(lr){1-6}
\textit{Average}
&                           & 3.5879 & 0.8170 & 0.8102 & 0.0071\\
\midrule

\multirow{3}{*}{Levant}
& Jordanian Arabic        & 5.0646 & 0.8146 & 0.8229 & 0.0143 \\
& Palestinian Arabic      & 2.4064 & 0.8173 & 0.8042 & 0.0000 \\
& Syrian Arabic           & 0.9218 & 0.8006 & 0.8063 & 0.0071 \\
\cmidrule(lr){1-6}
\textit{Average}
&                           & 2.7976 & 0.8108 & 0.8111 & 0.0071 \\
\midrule

\multirow{6}{*}{North Africa}
& Algerian Arabic         & 2.0994 & 0.8179 & 0.8107 & 0.0000 \\
& Egyptian Arabic         & 0.7131 & 0.8030 & 0.8168 & 0.0000 \\
& Libyan Arabic           & 1.3509 & 0.8037 & 0.7955 & 0.0000 \\
& Mauritanian Arabic      & 0.3946 & 0.8009 & 0.7922 & 0.0000 \\
& Moroccan Arabic         & 4.9470 & 0.8030 & 0.8060 & 0.0000 \\
& Tunisian Arabic         & 0.4923 & 0.8086 & 0.8017 & 0.0000 \\
\cmidrule(lr){1-6}
\textit{Average}
&                           & 1.6662 & 0.8062 & 0.8038 & 0.0000 \\
\midrule

\multirow{9}{*}{Europe and USA}
& Catalan                 & 7.2655 & 0.8544 & 0.8452 & 0.0143 \\
& Danish                  & 1.6260 & 0.8197 & 0.8100 & 0.0000 \\
& English                 & 3.9562 & 0.8470 & 0.8276 & 0.0000 \\
& French                  & 10.5773 & 0.8453 & 0.8321 & 0.0000 \\
& German                  & 1.0582 & 0.8249 & 0.8215 & 0.0000 \\
& Italian                 & 8.1754 & 0.8374 & 0.8321 & 0.0214 \\
& Luxembourgish           & 4.9795 & 0.8212 & 0.8182 & 0.0000 \\
& Portuguese              & 4.9488 & 0.8437 & 0.8349 & 0.0286 \\
& Spanish                 & 9.2473 & 0.8334 & 0.8194 & 0.0143 \\
\cmidrule(lr){1-6}
\textit{Average}
&                         & \textbf{5.7593} & \textbf{0.8363} & \textbf{0.8268} & 0.0087 \\
\midrule

\multirow{3}{*}{South Asia}
& Mandarin                & 10.3677 & 0.8284 & 0.8254 & 0.0500 \\
& Hindi                   & 0.4419 & 0.7950 & 0.7997 & 0.0000 \\
& Kashmiri                & 0.7335 & 0.7961 & 0.7874 & 0.0000 \\
\cmidrule(lr){1-6}
\textit{Average}
&                           & 3.8477 & 0.8065 & 0.8042 & 0.0167 \\
\bottomrule

\end{tabular}
\caption{Mean models performance on Task 3 grouped by language and region, measured using BLEU, Semantic Similarity (SS), BERTScore, and Exact Match (EM).}
\label{tab:task3-regions}

\end{threeparttable}
\end{table*}

\section{Selected Examples from \textsc{WISDOMBridge}}
\label{app:examples}

Table~\ref{tab:test-2-examples} shows examples from \texttt{Test\_2}. The pairs illustrate the intended evaluation behavior: the target proverb should be a recognized expression in the target language that conveys the same pragmatic lesson as the source proverb, even when the literal imagery differs.

\begin{table*}[t]
\centering
\scriptsize
\setlength{\tabcolsep}{6pt}

\begin{threeparttable}

\caption{Example values from Test\_2 showing proverb translation pairs across languages.}
\label{tab:test-2-examples}

\begin{tabular}{llll}
\toprule
\textbf{Source Language} &
\textbf{Source Proverb} &
\textbf{Target Language} &
\textbf{Target Proverb} \\
\midrule

English
& Bad news travels fast.
& Italian
& Cattive nuove presto corrono \\

Moroccan Arabic
& \ar{ارض بالقليل يعطيك الله لكثير}
& Portuguese
& Cada hum se contente com o que Deos lhe dá \\

Portuguese
& Boca de mel, coração de fel
& Saudi Arabic
& \ar{لسان عسل و فعل حنظل} \\

Mandarin
& \begin{chinese}混水里，好拿鱼\end{chinese}
& English
& It is good fishing in troubled waters \\

Egyptian Arabic
& \ar{إسأل مجرب ولا تسأل طبيب}
& Moroccan Arabic
& \ar{سل لمجرب، لا تسل الطبيب} \\

\bottomrule
\end{tabular}

\end{threeparttable}
\end{table*}

\section{Qualitative Error Analysis}
\label{app:error-analysis}

This section presents a qualitative error analysis of model outputs for the proverb explanation and proverb equivalence tasks. Through selected success and failure cases, we examine the strengths and limitations of the models, focusing on their ability to capture figurative meanings, interpret culturally specific expressions, and identify semantic equivalences across languages and cultures.

\begin{figure*}[htp] 
    \centering
    \subfloat[Proverb explanation]{%
        \includegraphics[width=0.45\textwidth]{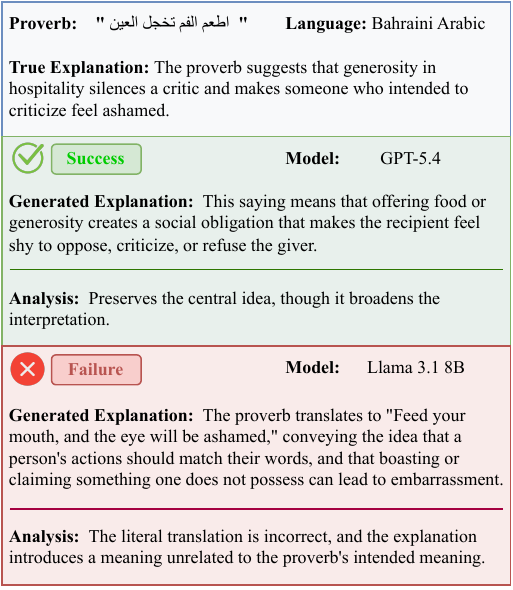}%
        \label{fig:explanation-error}%
        }%
    \hfill%
    \subfloat[Proverb Equivalence]{%
        \includegraphics[width=0.45\textwidth]{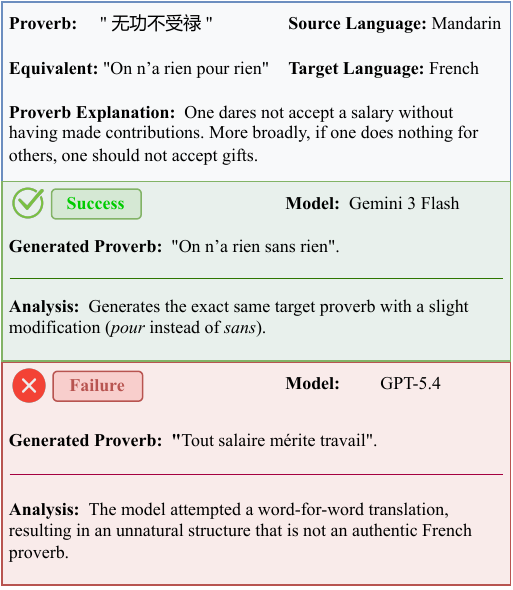}%
        \label{fig:equivalence-error}%
        }%
    \caption{Error analysis of the proverb explanation and proverb equivalence tasks, illustrating successful and failed model predictions alongside qualitative analyses of their outputs.}
\end{figure*}

Figure \ref{fig:explanation-error} illustrates successful and failed model-generated proverb explanations, highlighting issues such as misinterpretation of figurative meaning, and inaccurate literal translations, while Figure \ref{fig:equivalence-error} showcase successful and failed proverb equivalence predictions, highlighting challenges in identifying semantically equivalent proverbs across languages and cultures.

\end{document}